\documentclass[conference]{IEEEtran}
\IEEEoverridecommandlockouts
\usepackage{cite}
\usepackage{amsmath,amssymb,amsfonts}
\usepackage{algorithmic}
\usepackage{graphicx}
\usepackage{textcomp}
\usepackage{float}

\usepackage{xcolor}
\def\BibTeX{{\rm B\kern-.05em{\sc i\kern-.025em b}\kern-.08em
    T\kern-.1667em\lower.7ex\hbox{E}\kern-.125emX}}
\begin{document}

\title{Domain-Adaptive Learning: Unsupervised Adaptation for Histology Images with Improved Loss Function Combination\\

}

\author{\IEEEauthorblockN{1\textsuperscript{st} Ravi Kant Gupta}
\IEEEauthorblockA{\textit{Department of Electrical Engineering} \\
\textit{Indian Institute of Technology }\\
Mumbai, India \\
ravigupta131@gmail.com}
\and
\IEEEauthorblockN{2\textsuperscript{nd} Shounak Das}
\IEEEauthorblockA{\textit{Department of Electrical Engineering} \\
\textit{Indian Institute of Technology}\\
Mumbai, India \\
shounakd56@gmail.com}
\and
\IEEEauthorblockN{3\textsuperscript{rd} Amit Sethi}
\IEEEauthorblockA{\textit{Department of Electrical Engineering} \\
\textit{Indian Institute of Technology}\\
Mumbai, India \\
asethi@iitb.ac.in}
}

\maketitle

\begin{abstract}
This paper presents a novel approach for unsupervised domain adaptation (UDA) targeting H\&E stained histology images. Existing adversarial domain adaptation methods may not effectively align different domains of multimodal distributions associated with classification problems. The objective is to enhance domain alignment and reduce domain shifts between these domains by leveraging their unique characteristics. Our approach proposes a novel loss function along with carefully selected existing loss functions tailored to address the challenges specific to histology images. This loss combination not only makes the model accurate and robust but also faster in terms of training convergence. We specifically focus on leveraging histology-specific features, such as tissue structure and cell morphology, to enhance adaptation performance in the histology domain. The proposed method is extensively evaluated in accuracy, robustness, and generalization, surpassing state-of-the-art techniques for histology images. We conducted extensive experiments on the FHIST dataset and the results show that our proposed method - Domain Adaptive Learning (DAL) significantly surpasses the ViT-based and CNN-based SoTA methods by 1.41\%  and 6.56\% respectively.
\end{abstract}


\section{Introduction}
In traditional supervised learning, a model is trained using labeled data from the same domain as the test data. Obtaining labels for medical data is challenging due to the intricacies of medical expertise, making it costly and time-consuming. The need for specialized knowledge, meticulous review, and ethical considerations contribute to the difficulty in acquiring accurate and reliable annotations for medical datasets. However, when the distribution of the source and target domains differs significantly, the model's performance may suffer due to the domain shift. This domain shift can be because of color variation, data acquisition bias, distributional differences, domain-specific factors, covariate shift, staining techniques in medical histology images, etc. Unsupervised domain adaptation (UDA) techniques aim to mitigate this domain shift by aligning the feature distributions or learning domain-invariant representations by using only unlabeled samples from the target domain. Adversarial-based UDA employs a domain adversarial training framework, often based on Generative Adversarial Networks (GANs)~\cite{b32} or domain adversarial neural networks (DANN)~\cite{b1}. By learning domain-invariant representations, adversarial-based UDA models can effectively reduce the domain discrepancy and improve the generalization performance on the target domain. This approach has shown promising results in various domains, such as image classification, object detection, and semantic segmentation. However, while adversarial-based UDA has achieved notable success, challenges still exist. These include addressing the sensitivity to hyper-parameter tuning, handling the high-dimensional feature space, and effectively capturing complex domain shifts. 

\begin{figure}[!]
\centering
\includegraphics[height=5.5cm,width=7.5cm]{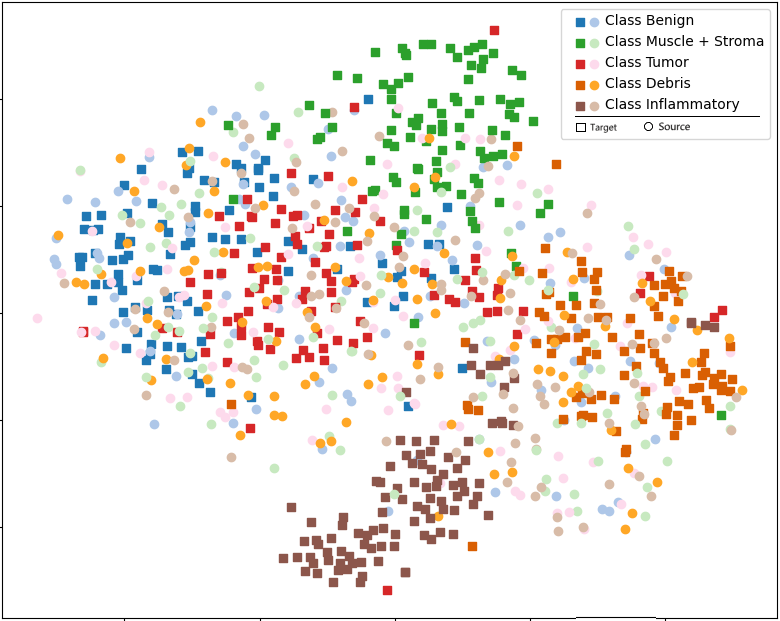}
\caption{Snapshot of t-SNE plot of source (CRC-TP) (Circle shape) and target (NCT) (Square shape), clearly shows significant difference between source and target data distribution.}
\label{st_tsne}
\end{figure}

\begin{figure*} 
\centering
\includegraphics[height=5cm,width=16cm]{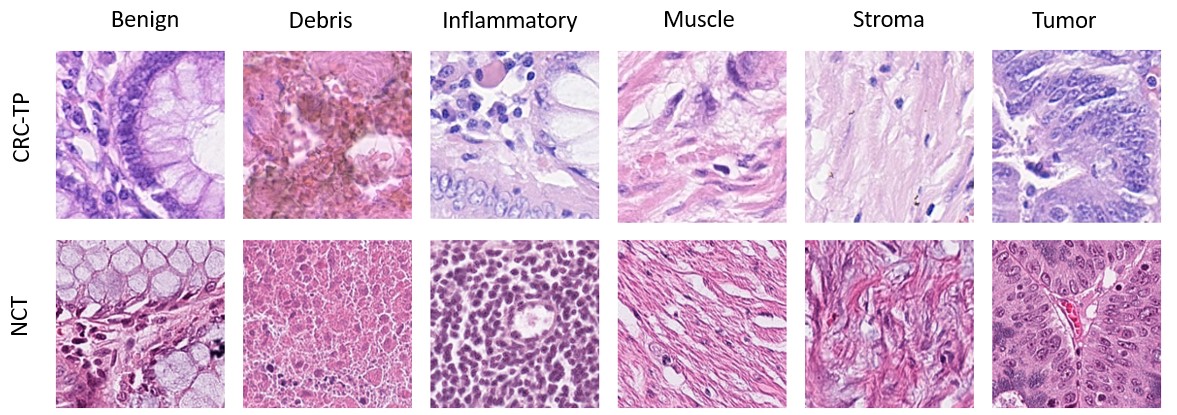}
\caption{Snapshot of sample images of each class from CRC-TP (top row) and NCT (bottom row) of FHIST dataset.}
\label{fig1}
\end{figure*}

To address the aforementioned challenge, we develop an unsupervised domain adaptation approach that surpasses the state-of-the-art performance for histological images. We present our findings from developing convolution neural networks (CNNs) for such tasks based on CRCTP dataset (source) and NCT (target) from FHIST dataset~\cite{b58}, which is composed of several histology datasets, namely CRC-TP~\cite{b41}, LC25000~\cite{b42}, BreakHis~\cite{b43}, and NCT-CRC-HE-100K~\cite{b44}.  These histology datasets consist of different tissue types and different organs. We consider each tissue type as a class label with one-hot encoding in the classification task. We framed our experiments on CRCTP and NCT with six classes (Benign, Tumor, Muscle, Stroma, Debris, and Inflammatory).  The t-distributed stochastic neighbor embedding (tSNE)~\cite{b65} plot in Figure \ref{st_tsne} of source data distribution (circle shape) and target data distribution (square shape) while the color of classes differs with light and dark versions of the same color. The sample patches of each class from both domains are shown in Figure \ref{fig1}.

Our research endeavors converge on three key objectives: firstly, to reduce the discordance between source and target domains in histology images; secondly, to harness the distinctive attributes of histology, such as cellular morphology and tissue structure, to elevate adaptation performance specifically within the histology domain; and finally, to transcend the limitations of current UDA techniques, to achieve state-of-the-art accuracy, resilience, and generalization capabilities compared to the previous methods. This comprehensive approach is crafted to encompass and intricately tackle the complexities presented by histology images. 

Our adoption of deep learning for unsupervised domain adaptation in histology images is driven by its potential to enhance model generalization, extract optimal features, enable versatile cross-domain applications, and achieve field-advancing progress. By tailoring the combination of loss functions which leads to improved convergence and robustness, and with the leverage of deep learning's power, we aim to surpass current methods, benefiting various applications. Inspired by a conditional domain adversarial network (CDAN)~~\cite{b3}, the core idea is to simultaneously train a feature extractor (typically a deep neural network) and a domain classifier (discriminator) to distinguish between source and target domains.  We have examined different CNN-based feature extractor as ResNet-50~\cite{b2}, ResNet-101~\cite{b2}, ResNet-152~\cite{b2}, VIT~\cite{b55}, and ConvMixer~\cite{b57} to extract meaningful features. The feature extractor aims to learn domain-invariant representations, while the domain classifier tries to classify the domain of the extracted features correctly. During training, the feature extractor and domain classifier are optimized in an adversarial manner. The feature extractor aims to fool the domain classifier by generating indistinguishable features across domains, while the domain classifier tries to classify the domains correctly. This adversarial training process encourages the feature extractor to learn domain-invariant and transferable representations between the source and target domains. To achieve this, we propose a novel loss function pseudo label maximum mean discrepancy (PLMMD) along with different existing losses such as maximum information loss (entropy loss)~\cite{b52,b53}, maximum mean discrepancy (MMD) loss~\cite{b14,b15,b16,b37}, minimum class confusion (MCC) loss~\cite{b54}, etc. This combination of loss functions has the following specific advantages :
Employing MCC loss enhances classification models by minimizing class confusion, particularly in scenarios with imbalanced class distributions. With maximum information loss, our model is encouraged to learn tightly clustered target features with uniform distribution, such that the discriminative information in the target domain is retained, while MDD loss measures the difference between the mean embeddings of two distributions, helping to quantify the dissimilarity between domains and facilitating domain adaptation techniques. Our proposed loss PLMMD enhances unsupervised domain adaptation by selectively emphasizing domain-invariant features through weight assignments. The benefit of this novel loss is, that training convergence is faster as compared to other scenarios. With the help of this novel combination of the loss function our method surpasses not only the CNN-based model state-of-art but also the transformer-based model for the histology images. To justify our claims for histology images, we use the FHIST dataset~\cite{b58}, whose sample images are shown in Figure~\ref{fig1}.

Our stated goals were achieved by proposing an improved combination of loss functions tailored to address the unique challenges of H\&E stained histology images. The performance evaluation was focused on accuracy, robustness, and generalization, to surpass state-of-the-art techniques in both domains. Furthermore, the research explored potential cross-domain applications in medical image analysis and computer vision, offering promising advancements in practical unsupervised domain adaptation with the help of various combinations of loss functions with different existing models.

\section{Background and Related Work}

In unsupervised domain adaptation, we have a source domain $D_s = \{(x_{s_i}, y_{s_i})\}_{i=1}^{n_s}$ of ${n_s}$ labeled examples and a target domain $D_s = \{(x_{t_i}, y_{t_i})\}_{i=1}^{n_t}$ of ${n_t}$ unlabeled examples. The source domain and target domain are sampled from joint distributions $P({x_s},{y_s})$ and $Q({x_t},{y_t})$ respectively. Notably, the two distributions are initially not aligned, that is, $P \neq Q$.

Domain adversarial neural network (DANN)~\cite{b1} is a framework of choice for UDA. It is a two-player game between domain discriminator $D$, which is trained to distinguish the source domain from the target domain, and the feature representation $F$ trained to confuse the domain discriminator $D$ as well as classify the source domain samples. The error function of the domain discriminator corresponds well to the discrepancy between the feature
distributions $P(f)$ and $Q(f)$~\cite{b51}, a key to bound the target risk in the domain adaptation theory~\cite{b66}.

Alignment-based domain adaptation is another typical line of work that leverages a domain-adversarial task
to align the source and target domains as a whole so that class labels can be transferred from the source domain to
the unlabeled target one~\cite{b10,b11,b12,b13}. Another typical line of work directly minimizes the domain shift measured by various metrics, e.g., maximum mean discrepancy (MMD)~\cite{b14,b15,b16}. These methods are based on domain-level domain alignment. To achieve class-level domain alignment, the works of~\cite{b17,b18} utilize the multiplicative interaction of feature representations and class predictions so that the domain discriminator can be aware of the classification boundary.
Based on the integrated task and domain classifier,~\cite{b19} encourages a mutually inhibitory relation between category and domain predictions for any input instance. The works of~\cite{b20, b23} align the labeled source centroid and pseudo-labeled target centroid of each shared class in the feature space. Some work uses individual task classifiers for the two domains to detect non-discriminative features and reversely learn a discriminative feature extractor~\cite{b24, b25, b26}. Certain other works focus attention on transferable regions to derive a domain-invariant classification model~\cite{b27,b28,b29}. To help achieve target-discriminative features,~\cite{b30,b31} generate synthetic images from the raw input data of the two domains via GANs~\cite{b32}. The recent work of~\cite{b33} improves adversarial feature adaptation, where the discriminative structures of target data may be deteriorated~\cite{b34}. The work of~\cite{b35} adapts the feature norms of the two domains to a large range of values so that the learned features are not only task-discriminative but also domain-invariant.

\section{Proposed Method}
The challenge of domain shift in a cross-domain classification task using unsupervised domain adaptation leverages the knowledge from a labeled source domain to improve the performance of a classifier on an unlabeled target domain. We propose a novel loss function that minimizes the domain discrepancy and aligns feature distributions across domains. Our datasets even differ in patch sizes - 150x150 pixels for the source domain and 224x224 for the target domain. Before training, the patches were subjected to data augmentation such as horizontal flip, vertical flip, and normalization to ensure consistency. To facilitate domain adaptation, we introduce a structure-preserving colour normalization technique to normalize the stain appearance of histopathology images across domains~\cite{b56}. The normalization process aims to preserve the local structure while removing domain-specific variations. Therefore, the patches were colour normalized~\cite{b38,b39}.

\begin{figure*}[!]
\centering
\includegraphics[height=6cm]{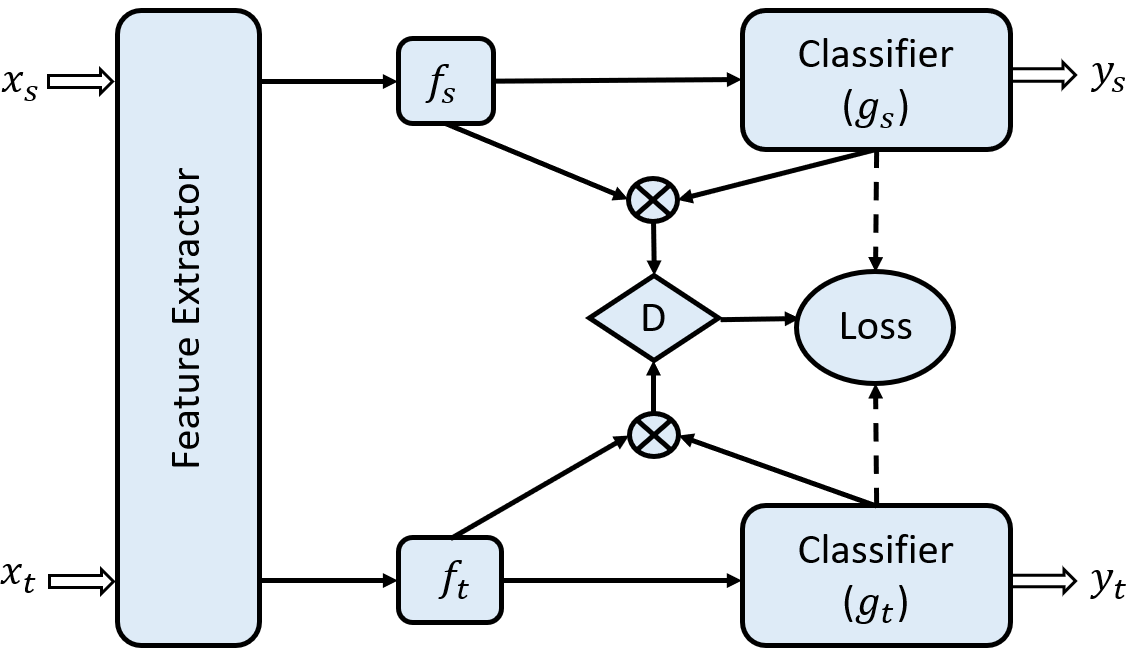}
\caption{Architecture of the proposed networks, where domain-specific feature representation $f$ and classifier prediction $g$ embody the cross-domain gap to be reduced jointly by the conditional domain discriminator $D$.}
\label{fig2}
\end{figure*}

From the color-normalized patches, we extracted features using ResNet152 trained on ImageNet~\cite{b40}. Our proposed model architecture is based on a deep neural network with convolutional and fully connected layers, specifically tailored for domain adaptation.

 In this work, we design a method to train a deep network $N: x \rightarrow y$ which reduces the shifts in the data distributions across domains, such that the target risk $r_t$= \(E_{(x_t, y_t) \sim Q}[N(x_t) \neq y_t]\) can be bounded by the source risk $r_s$= \(E_{(x_s, y_s) \sim P}[N(x_s) \neq y_s]\) plus the distribution discrepancy disc(P, Q) quantified by a novel conditional domain discriminator. To minimize domain cross-domain discrepancy~\cite{b1} in adversarial learning Generative Adversarial Networks (GANs)~\cite{} play a vital role. Features are represented by $f = F (x)$ and classifier prediction, $g = N(x)$ generated from deep network $N$.

We improve existing adversarial domain adaptation methods in two directions. First, when the joint distributions of feature and class, i.e. $P({x_s},{y_s})$ and $Q({x_t},{y_t})$, are non-identical across domains, adapting only the feature representation $f$ may be insufficient. A quantitative study~\cite{b59} shows that deep representations eventually transition from general to specific along deep networks, with transferability decreased remarkably in the domain-specific feature layer $f$ and classifier layer $g$.
Second, due to the nature of multi-class classification, the feature distribution is multimodal, and hence adapting feature distribution may be challenging for adversarial networks.

By conditioning, domain variances in feature representation $f$ and classifier prediction $g$ can be modeled simultaneously. This joint conditioning allows us to bridge the domain gap more effectively, enabling the adapted model to capture and align the underlying data distributions between the source and target domains. Consequently, incorporating classifier prediction as a conditioning factor in domain adaptation holds great potential for achieving improved transferability and generating domain-invariant representations in challenging cross-domain scenarios.

We formulate Conditional Domain Adversarial Network (CDAN)~\cite{b3} as a minimax optimization problem with two competitive error terms: (a) $E(N)$ on the source classifier N, which is minimized to guarantee lower source risk; (b) $E(D, N)$ on the source classifier $N$ and the domain discriminator $D$ across the source and target domains, which is minimized over $D$ but maximized over $f = F(x)$ and $g = N(x)$:
\begin{equation}
 L_{clc}(x_{s_i}, y_{s_i}) = \mathbb{E}_{(x_{s_i}, y_{s_i}) \sim D_s} L(N(x_{s_i}), y_{s_i}) 
\label{eqn1}
\end{equation}

\begin{equation}
\begin{aligned}
L_{dis}(x_s, x_t) = & -\mathbb{E}_{x_{s_i} \sim D_s} \log [D(f_{s_i}, g_{s_i})] \\
         & -\mathbb{E}_{x_{t_j} \sim D_t} \log [1 - D(f_{t_j}, g_{t_j})],
\end{aligned}
\label{eqn2}
\end{equation}

where $L$ is the cross-entropy loss, and $h = (f, g)$ is the joint variable of feature representation $f$
and classifier prediction $g$. The minimax game of CDAN is
\begin{equation}
\begin{aligned}
\begin{gathered}
\min_{N} L_{clc}(x_{s_i}, y_{s_i}) - \lambda L_{dis}(x_s, x_t) \\
\min_{D} L_{dis}(x_s, x_t),
\end{gathered}
\end{aligned}
\label{eqn3}
\end{equation}
where $\lambda$ is a hyper-parameter between the two objectives to trade off source risk and domain adversary.

We condition domain discriminator $D$ on the classifier prediction g through joint variable $h = (f, g)$ to potentially tackle the two aforementioned challenges of adversarial domain adaptation. A simple conditioning of $D$ is \(D(f \oplus g)\), where we concatenate the feature representation and classifier prediction in vector \(f \oplus g\) and feed it to conditional domain discriminator $D$. This conditioning strategy is widely adopted by existing conditional GANs~\cite{b32}. However, with the concatenation strategy, $f$ and $g$ are independent of each other, thus failing to fully capture multiplicative interactions between feature representation and classifier prediction, which are crucial to domain adaptation. As a result, the multimodal information conveyed in classifier prediction cannot be fully exploited to match the multimodal distributions of complex domains~\cite{b60}. The multilinear map is defined as the outer product of multiple random vectors. And the multilinear map of infinite-dimensional nonlinear feature maps has been successfully applied to embed joint distribution or conditional distribution into reproducing kernel Hilbert spaces~\cite{b60,b61,b62,b63}. 
Besides the theoretical benefit of the multilinear map \(x \otimes y\) over the concatenation \(x \oplus y\)~\cite{b60,b64}.  Taking advantage of the multilinear map, in this paper, we condition $D$ on $g$ with the multilinear map. Superior to concatenation, the multilinear map \(x \otimes y\) can fully capture the multimodal structures behind complex data distributions. A disadvantage of the multilinear map is dimension explosion.

We enable conditional adversarial domain adaptation over domain-specific feature representation $f$ and classifier prediction $g$. We jointly minimize with respect to (\ref{eqn1})  source classifier $N$ and feature extractor $F$, minimize (\ref{eqn2})  domain discriminator $D$, and maximize (\ref{eqn2})  feature extractor $F$ and source
classifier $N$. This yields the mini-max problem of Domain Adversarial Networks:
\begin{align}
\min_{G} \quad & \mathbb{E}_{(x_s^i, y_s^i)\sim D_s} L(G(x_s^i), y_s^i) \notag \\
& + \lambda\left( \mathbb{E}_{x_s^i \sim D_s}\log [D(T(h_s^i))] \right. \notag \\
& \left. + \mathbb{E}_{x_t^j \sim D_t}\log[1 - D(T(h_t^j))]\right) \\
\max_{D} \quad & \mathbb{E}_{x_s^i \sim D_s} \log [D(T(h_s^i))] + \mathbb{E}_{x_t^j \sim D_t} \log [1 - D(T(h_t^j))],
\end{align}

where $\lambda$ is a hyper-parameter between the source classifier and conditional domain discriminator, and
note that $h = (f, g)$ is the joint variable of domain-specific feature representation $f$ and classifier
prediction $g$ for adversarial adaptation.

The general problem of adversarial domain adaptation of the proposed model for classification can be formulated as follows:

\begin{equation}
\begin{aligned}
L=\min_{N} L_{clc}(x_{s_i}, y_{s_i}) - \lambda L_{dis}(x_s, x_t)\\ +\beta L_{IM} + \gamma L_{MCC} + \delta L_{MDD} +  \eta L_{WMMD} 
\end{aligned}
\end{equation}

where $\lambda$, $\beta$, $\gamma$, $\delta$ and $\eta$ are hyper parameters, $L_{MCC}$ is minimum class confusion loss, $L_{MDD}$ is maximum mean discrepancy loss, $L_{WMDD}$ represents weighted maximum mean discrepancy loss and $L_{IM}$ represents information maximization loss. All individual losses have their own specialty and this novel combination of loss significantly surpasses the performance of CNN-based models as well as transformer-based models. A detailed description of all the losses is given below in the losses section.

\subsection{Losses}
\subsubsection{ Maximum Mean Discrepancy}
Maximum mean discrepancy (MMD) is a kernel-based statistical test used to determine whether given two distributions are the same~\cite{b14,b15,b16}.
Given an random variable {X} , a feature map $\phi$ maps {X} to an another space $F$ such that $\phi(X) \in {F}$
.
Assuming $F$ satisfies the necessary conditions, we can benefit from the kernel trick to compute the inner product in $F$ :
\begin{equation}
     X, Y \text{ such that } k(X,Y) = \langle \phi(X), \phi(Y)\rangle _{F},
\end{equation}
where $k$ is gram matrix produced using the kernel function.
\\
MMD is the distance between feature means. That means for a given probability measure $P$ on $X$, feature means is an another feature map that takes $\phi(X)$ and maps it to the means of every coordinate of $\phi(X)$ :
\begin{equation}
\mu_{p}(\phi(X)) = [\mathbb{E}[\phi(X_{1})], ...., \mathbb{E}[\phi(X_{m})]]^T
\end{equation}
The inner product of feature means of $X \sim P$ and $Y\sim Q$ can be written in terms of kernel function such that:
\begin{equation}
\begin{aligned}
\begin{gathered}
\langle \mu_{p}(\phi(X)),\mu_{q}(\phi(Y)) \rangle_{F} = \mathbb{E}_{P,Q}[\langle \phi(X),\phi(Y) \rangle_{F}]\\=\mathbb{E}_{P,Q}[k(X,Y)]
\end{gathered}
\end{aligned}
\label{eqn_}
\end{equation}
Given $X$, $Y$ maximum mean discrepancy is the distance between feature means of $X$, $Y$ :
\begin{equation}
MMD^2 (P, Q) = ||\mu_{P}-\mu_{Q}||^2 _{F}
\end{equation}

\begin{equation}
\begin{aligned}
\begin{gathered}
MMD^2 (P, Q) = \langle \mu_{P}-\mu_{Q}, \mu_{P}-\mu_{Q} \rangle \\
= \langle \mu_{P}, \mu_{P}\rangle - 2\langle \mu_{P}, \mu_{Q}\rangle + \langle \mu_{Q}, \mu_{Q}\rangle 
\end{gathered}
\end{aligned}
\end{equation}
Using the equation (\ref{eqn_}), finally above expression becomes
\begin{equation}
\begin{aligned}
\begin{gathered}
{L}_{MMD}=MMD^2 (P, Q) \\= \mathbb{E}_{P}[k(X,X)] - 2\mathbb{E}_{P,Q}[k(X,Y)] +\mathbb{E}_{Q}[k(Y,Y)]
\end{gathered}
\end{aligned}
\label{eqn11}
\end{equation}

\subsubsection{Pseudo Label Maximum Mean Discrepancy}
We calculated the PLMMD using a similar procedure to calculating MMD loss in equation (\ref{eqn11}). However, our proposed loss differs in terms of weights assigned to each similarity term. Hence we can define PLMMD loss as:

\begin{equation}
\begin{aligned} 
\begin{gathered}
{L}_{PLMMD}= w_{XX} \mathbb{E}_{P}[k(X,X)] - 2 w_{XY} \mathbb{E}_{P,Q}[k(X,Y)] \\+ w_{YY}\mathbb{E}_{Q}[k(Y,Y)],
\end{gathered}
\end{aligned}
\end{equation}
where, $w_{XX}$ represent weight to get similarity within the source domain, similarly, $w_{YY}$ are weights for similarity within the target domain, and $w_{XY}$ are weights to get similarity within source and target domain. For calculating the weights, first, we generated pseudo labels for the target using a source classifier. After that, the source and target pseudo-label is normalized to account for class imbalances. For each class common to both datasets, dot products of normalized vectors are computed to quantify instance relationships. Calculated dot products are normalized by the count of common classes, ensuring fairness. This returns three weight arrays, representing relationships between instances in the source dataset, target dataset, and source-to-target pairs.

\subsubsection{Minimum Class Confusion} 
The minimum class confusion loss $\mathcal{L}_{MCC}$~~\cite{b54} seeks to minimize confusion terms between classes $j$ and $j'$, such that $j \neq j'$ where the indices are exhaustive over the set of classes. On the target domain, the class confusion term between two classes $j$ and $j'$ is given by:  \begin{center}
     $C_{jj'} =  \hat{\mathbf{y}}_{\cdot j}^{\intercal} \hat{\mathbf{y}}_{\cdot j'}^{\intercal}$ 
 \end{center} 
 
A much more nuanced and meaningful formulation of the class confusion would be:

 \begin{equation}
     C_{jj'} = \hat{\mathbf{y}}_{\cdot j}^{\intercal} \mathbf{W} \hat{\mathbf{y}}_{\cdot j'}^{\intercal},
 \end{equation}
where the matrix $\mathbf{W}$ is a diagonal matrix. The diagonal terms $W_{ii}$ are given as the softmax outputs of the entropies in classifying a sample $i$. $\hat{\mathbf{y}}_{ij}$ is given as: 
  
 \begin{equation}
  \hat{\mathbf{y}}_{ij} = \frac{ \exp(Z_{ij} / T ) }{\sum_{j'=1} ^ {c} \exp (Z_{ij'}/ T )} ,   
 \end{equation}
where $c$ is the number of classes, $T$ is the temperature coefficient, and $Z_{ij}$ is the logistic output of the classifier layer for the class $j$ and the sample $i$. 

After normalizing the class confusion terms, the final MCC Loss function is given as:
 \begin{equation}
     \mathcal{L}_{MCC} = \frac{1}{c} \sum_{j=1}^ {c} \sum_{j' \neq j}^ {c}|C_{jj'}|,
 \end{equation}
which is the sum of all the non-diagonal elements of the class confusion matrix. The diagonal terms represent the "certainty" in the classifier, while the non-diagonal terms represent the "uncertainty" in classification. The MCC loss can be added in conjunction with other domain adaptation methods.

\subsubsection{Information Maximization loss}

The Information Maximization loss is designed to encourage neural networks to learn more informative representations by maximizing the mutual information between the learned features and the input data~\cite{b52,b53}. This type of loss aims to guide the model to capture relevant and distinctive patterns in the data, which can be especially valuable in scenarios where unsupervised learning, domain adaptation, or feature learning are important. The assumptions that \(p_t = \text{softmax}(N(f(x_t)))\) are expected to retain as much information about $x_t$ as possible, and decision
boundary should not cross high-density regions, but instead
lie in low-density regions, which is also known as cluster
assumption. These two assumptions can be met by maximizing mutual information between the empirical distribution of the target inputs and the induced target label distribution, which can
be formally defined as:

\begin{equation}
\begin{gathered}
I(p_t; x_t) = H(\overline{p}_t) - \frac{1}{n_t} \sum_{j=1}^{n_t} H(p_{tj}) \\
= -\sum_{k=1}^{K} \overline{p}_{tk} \log(\overline{p}_{tk}) + \frac{1}{n_t} \sum_{j=1}^{n_t} \sum_{k=1}^{K} p_{tkj} \log(p_{tkj}),
\end{gathered}
\end{equation}
where, \(p_{tj} = \text{softmax}(G_c(G_f(x_{tj})))\), \(\overline{p}_t = \mathbb{E}_{x_t}[p_t]\), and K is the number of classes. Maximizing $-\frac{1}{n_t} \sum_{j=1}^{n_t} H(p_{tj})$ enforces the target predictions close to one-hot encoding, therefore the cluster assumption is guaranteed. To ensure global diversity, we also maximize \(H(\overline{p}_t)\) to avoid every target data being assigned to the same class.
With \(I(p_t; x_t)\), our model is encouraged to learn tightly clustered target features with uniform distribution, such that the discriminative information in the target domain is retained. 

\begin{figure*}[!]
\centering
\includegraphics[height=4.5cm,width=16cm]{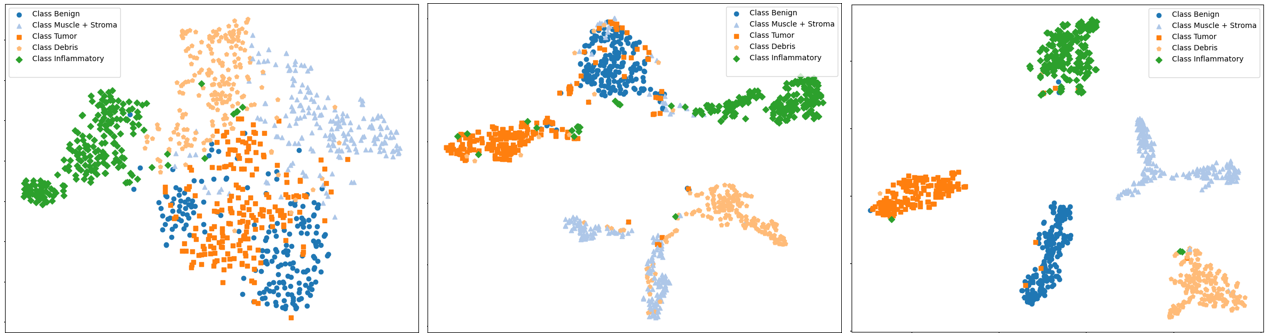}
\caption{Snapshots of 2D tSNE plots of the target (NCT) domain sample features before training (leftmost), after three epochs (middle), and after six epochs (right).}
\label{tsne}
\end{figure*}

\section{Experimentation and Results}

\textbf{Dataset:} To evaluate the proposed method, we introduce the FHIST dataset, a proposed benchmark for the few-shot classification of histological images~\cite{b58}. FHIST is composed of several histology datasets, namely CRC-TP~\cite{b41}, LC25000~\cite{b42}, BreakHis~\cite{b43}, and NCT-CRC-HE-100K~\cite{b44}. For each class, there are close to 20,000 patches in the CRC-TP domain with a patch size of 150X150 pixels and around 10,000 patches of size 224X224 pixels in the NCT domain. We performed experiments with CRC-TP as the source and NCT as the target and vice versa. The tSNE plots shown in Figure \ref{tsne} depict the distribution of target (NCT) at different stages of training. Different colors map different class types in the tSNE plot. We have plotted five classes in tSNE which are Benign, Tumor, Debris, Inflammatory, and Muscle + Stroma with 200 sample points from each five classes. We combined the last two classes because of their physiological as well as feature intertwining.  The first plot(leftmost) shows the data distribution of NCT(as target) at epoch 0, and the second one shows the data distribution of NCT after four epochs, and the last one (rightmost) shows the target(NCT) data distribution after six epochs of domain adaptation. These histology datasets consist of different tissue types and different organs. We consider each tissue type as a class label with one-hot encoding in the classification task. We framed our experiments on CRC-TP and NCT with six classes (Benign, Tumor, Muscle, Stroma, Debris, and  Inflammatory)

\textbf{Implementation:} All the experiments were conducted on an NVIDIA A100 in PyTorch, using the CNN-based neural network (ResNet-152) pre-trained on ImageNet~\cite{b40} as the backbone for our proposed model. The base learning rate is 0.00001 with a batch size of 32, and we train models by 20 epochs. The hyper-parameters were $\beta$=0.05 , $\gamma$=1.4 , $\delta$=0.54 and $\eta$=0.54 for the experiment of CRC-TP $\rightarrow$ NCT and NCT $\rightarrow$ CRC-TP. We used AdamW~\cite{b45} with a momentum of 0.9, and a weight decay of 0.001 as the optimizer. We adopt the standard protocol for unsupervised domain adaptation (UDA) where all labeled source samples and unlabeled target samples are utilized for training. To report our results for each transfer task, we use center-crop images from the target domain and report the classification performance. For a fair comparison with prior works, we also conduct experiments with the same backbone as ViT-based~\cite{b55} as TVT~\cite{b48}, ResNet-50~\cite{b2}, DANN~\cite{b1}, CDAN~\cite{b3}, GVB-GD~\cite{b46}, CHATTY+MCC~\cite{b47} on FHIST dataset.

\begin{figure}
\centering
\includegraphics[height=6.5cm,width=9cm]{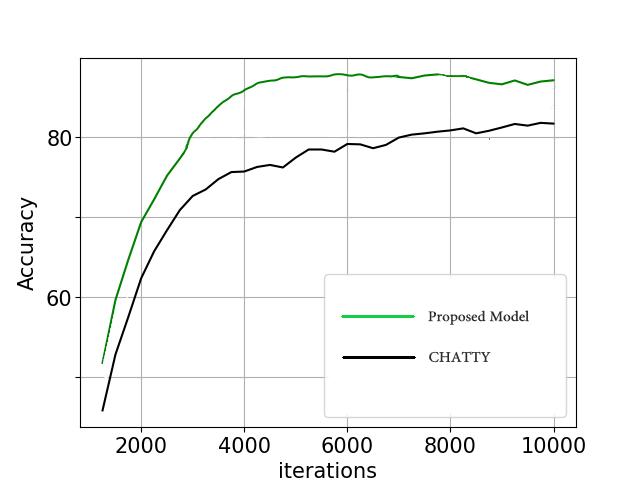}
\caption{Evolution of accuracy on FHIST dataset \cite{b58} CRC-TP to NCT domain adaptation task shows improvement in convergence with proposed model compare to CHATTY model \cite{b47}.}
\label{conv}
\end{figure}
\subsection{Dataset and Implementation}

\subsection{Results}
Our analysis in Table \ref{table1} depicts results with different methods and feature extractors for the FHIST dataset. The top five methods are CNN model using ResNet-50 as a feature extractor trained on ImageNet dataset while TVT uses ViT based model pretrained on ImageNet-21k dataset. Our proposed method is a CNN-based model that utilizes ResNet-152 as a backbone with a novel combination of loss functions. Our model outperforms CNN-based models such as ResNet-50, DANN, CDAN, GVB-GD, and CHATTY+MCC, and surpasses the state of the Art (SoTA) CNN results by 6.56\%. At the same time, our method also surpasses the transformer-based SoTA by 1.41\%. We achieved an accuracy of 87.71\% for CRC-TP to NCT domain adaptation and 74.81\%  for NCT to CRC-TP with an average accuracy of 81.26\%  for both tasks, as mentioned in Table \ref{table1} with bold text.

\begin{table*}
\caption{  Accuracy (\%) on the FHIST dataset \cite{b58} with two different UDA tasks and their average, where all methods are fine-tuned on their respective backbone model.}
\begin{center}
\begin{tabular}{|l|c|c|c|c|c|}
\hline
Method	& Backbone Feature Extractor	 &	CRC-TP $\rightarrow$ NCT	&	NCT $\rightarrow$ CRC-TP	&	Average	\\
\hline
ResNet-50~~\cite{b2}	&	 &	40.7	&	32.9	&	36.8	\\
DANN~~\cite{b1}	&	 &	73.5	&	66.6	&	70.0	\\
CDAN~~\cite{b3}	& ResNet-50 &	66.2	&	61.4	&	63.8	\\
GVB-GD~~\cite{b46}	&	 &	73.9	&	66.7	&	70.3	\\
CHATTY~\cite{b47}	&	 &	81.6	&	67.9	&	74.7	\\
\hline
TVT~\cite{b48} &	ViT &	86.47 & 73.23	&	79.85 \\
\hline
Proposed Method	&	ResNet-152 &	\textbf{87.71} & \textbf{74.81}	&	\textbf{81.26} \\
\hline
\end{tabular}
\end{center}  
\label{table1}
\end{table*}

\section*{Discussion and Conclusion}
In this study, we have demonstrated that utilizing different combinations of loss functions with a CNN such as ResNet-152 can lead to significant improvements in unsupervised domain adaptation (UDA) performance that can surpass the performance of ViTs using other UDA methods. By leveraging the strengths of various loss functions tailored to specific domain characteristics, we have surpassed the state-of-the-art (SOTA) performance for histology images. 
We conducted ablation studies to understand the impact of the different feature extractors such as ConvMixer~\cite{b57} and ResNet-101~\cite{b2}. However, the performance in these cases was worse than our reported results.
To know the impact of individual loss and a combination of losses, we performed extensive experiments. Through comprehensive experiments, we discovered that Minimum Class Confusion (MCC) loss functions offer an enhancement to classification models by mitigating class confusion, particularly when faced with imbalanced class distributions. In parallel, we observed that information maximization losses aid the classifier in selecting the most certain samples for domain alignment. In our proposed approach, the Pseudo Label Maximum Mean Discrepancy (PLMMD) accelerates training convergence (comparision with CHATTY model shown in Figure \ref{conv}) and notably enhances domain alignment by incorporating weighted considerations. Additionally, the Maximum Mean Discrepancy (MMD) loss effectively narrows the gap between the mean embeddings of the two distributions. By artfully combining these distinctive loss functions, we not only surpass the current state-of-the-art but also achieve a comprehensive solution that advances the field of classification models in diverse scenarios.



\end{document}